\documentclass[10pt,twocolumn,letterpaper]{article}
\usepackage{cvpr}              % To produce the CAMERA-READY version
\usepackage{xcolor}

\usepackage{pifont}
\usepackage{adjustbox}
\usepackage{multirow}
\usepackage{colortbl}
\usepackage{bm}

\newcommand{\fig}[2][1]{\includegraphics[draft=False, width=#1\linewidth]{#2}}
\definecolor{demphcolorinline}{gray}{.3}

\newcommand{\tablefirst}{\cellcolor{gray!10}}
\usepackage{algorithm}
\usepackage{algorithmic}

\usepackage{array}
\usepackage{tabularx}
\usepackage{makecell}
\usepackage{amsmath}
\usepackage{listings}
\usepackage{xcolor}
\usepackage[accsupp]{axessibility}  % Improves PDF readability for those with disabilities.
\definecolor{Green}{rgb}{0.0, 0.5, 0.0}

\definecolor{cvprblue}{rgb}{0.21,0.49,0.74}
\usepackage[pagebackref,breaklinks,colorlinks,allcolors=cvprblue]{hyperref}

\def\paperID{2325} % *** Enter the Paper ID here
\def\confName{CVPR}
\def\confYear{2025}

\title{Rethinking Query-based Transformer for Continual Image Segmentation}

\author{
Yuchen Zhu$^{1}$$^*$\hspace{1.5em}Cheng Shi$^{1}$\thanks{Equal contribution}\hspace{1.5em}Dingyou Wang$^{1}$\hspace{1.5em}Jiajin Tang$^{1}$\hspace{1.5em}Zhengxuan Wei$^{1}$\\\hspace{1.5em}Yu Wu$^{3}$\hspace{1.5em}Guanbin Li$^{2}$\hspace{1.5em}Sibei Yang$^{2}$\thanks{Corresponding author}\\
{$^1$School of Information Science and Technology, ShanghaiTech University}\\
{$^2$Sun Yat-sen University\hspace{1.5em}$^3$Wuhan University}\\
}

\begin{document}
\maketitle
\begin{abstract}
Class-incremental/Continual image segmentation (CIS) aims to train an image segmenter in stages, where the set of available categories differs at each stage. 
To leverage the built-in objectness of query-based transformers, which mitigates catastrophic forgetting of mask proposals, current methods often decouple mask generation from the continual learning process. 
This study, however, identifies two key issues with decoupled frameworks: loss of plasticity and heavy reliance on input data order. 
To address these, we conduct an in-depth investigation of the built-in objectness and find that highly aggregated image features provide a shortcut for queries to generate masks through simple feature alignment. 
Based on this, we propose SimCIS, a simple yet powerful baseline for CIS. Its core idea is to directly select image features for query assignment, ensuring ``perfect alignment" to preserve objectness, while simultaneously allowing queries to select new classes to promote plasticity. 
To further combat catastrophic forgetting of categories, we introduce cross-stage consistency in selection and an innovative ``visual query"-based replay mechanism. 
Experiments demonstrate that SimCIS consistently outperforms state-of-the-art methods across various segmentation tasks, settings, splits, and input data orders. 
All models and codes will be made publicly available at \url{https://github.com/SooLab/SimCIS}. 
\end{abstract}    
\section{Introduction}
\label{sec:intro}
\begin{figure}[t]
\begin{tabular}{cc}
\fig[.45]{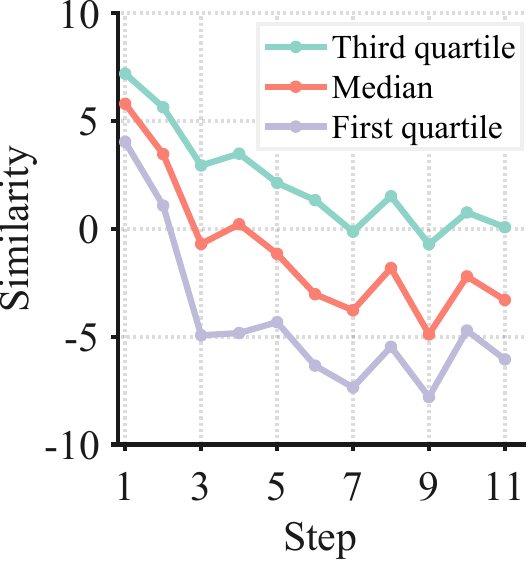} &
\fig[.46]{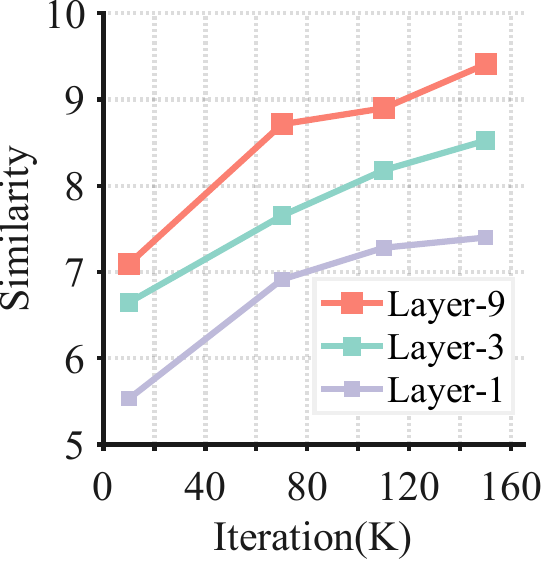} \\

% (a)& (b)  & (c)\\
\end{tabular}%
    \caption{
    \textbf{Similarity} between queries and feature map changes across decoder layers and training stages (right). The query gradually misaligns with the pixel feature (left).
    }
\label{fig:3}
\vspace{-6mm}
\end{figure}

Continual learning empowers models to progressively acquire, learn, and assimilate new knowledge from an ever-evolving environment. It serves as a fundamental task in image classification ~\cite{(rwil)chaudhry2018riemannian,(podnet)douillard2020podnet,(iCaRL)rebuffi2017icarl,(eteil)castro2018end,(packnet)mallya2018packnet,(slca)zhang2023slca,(der)yan2021dynamically,(cfrp)robins1995catastrophic,(cldgr)shin2017continual,shi2024part2object,dai2024curriculum,shi2024devil,tang2023temporal,lin2021structured, yang2021bottom, ge2018multi,he2019non} where models are required to recognize new classes (\textbf{plasticity}) and preserve old class knowledge (avoid \textbf{catastrophic forgetting}). Extending beyond classification, continual image segmentation adapts this to the image segmentation, unlocking a myriad of practical applications~\cite{(progress)rusu2016progressive,(inc)ross2008incremental}. However, it also confronts more challenges: 1) \textbf{Additional catastrophic forgetting} of mask prediction, beyond that of class prediction; 2) \textbf{Background semantic shift} occurs when the current foreground becomes background in subsequent stages, driven by the need for image segmentation to predict the background class and the constraint of only having class annotations from current stage.
\begin{figure}
    \centering
    \includegraphics[width=0.85\linewidth]{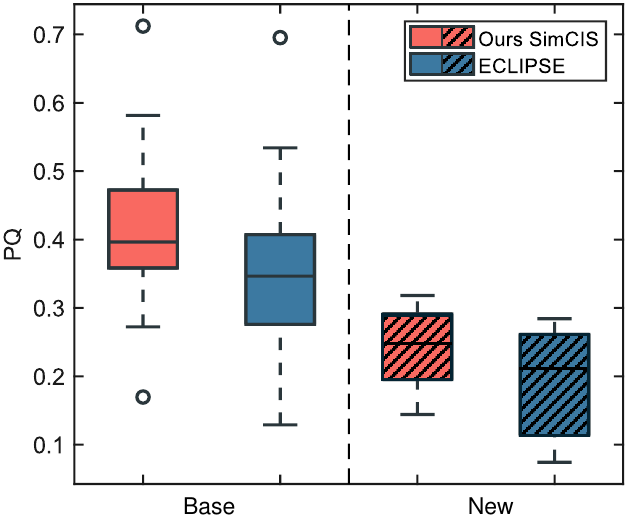}
    \caption{
    \textbf{Boxplots} of PQ metric for our SimCIS and previous SOTA~\cite{(eclipse)kim2024eclipse} on ADE20K. We train each model on randomly shuffled continual data input orders and report average PQ for base and novel classes. We observe that recent query-based transformers suffer from a loss of plasticity (low average PQ) and heavy reliance on the input data order (high variance). 
    }
    \label{fig:boxplots}
    \vspace{-10pt}
\end{figure}
Recently, query-based transformers~\cite{(mask2former)cheng2022masked,chen2024survey,tang2023contrastive,shi2024plain,huang2023free,zheng2023ddcot,huang2025mvtokenflow,shi2023logoprompt, shi2023edadet} are introduced into continual image segmentation, as their \textbf{built-in objectness} has been shown to mitigate catastrophic forgetting in mask generation. 
% As shown in [Fig:objectness], the query-based transformer retains its ability to generate mask proposals for old classes even after finetuning on new classes. 
Leveraging this built-in objectness, many studies~\cite{(eclipse)kim2024eclipse, (CoMasTRe)gong2024continual,(ssul)cha2021ssul,(RC)zhang2022representation} decouple mask segmentation from the continual learning process by freezing the parameters associated with mask proposal generation.
However, we observe two notable yet suboptimal behaviors in the aforementioned methods.
\begin{itemize}
    \item The advantage of objectness diminishes and even has a detrimental effect on plasticity as the task sequence shortens. In the shortest two-task setting, they typically achieve performance comparable to or even slightly lower than the baseline. % (Comformer~\cite{(comformer)cermelli2023comformer}).
    \item The built-in objectness is fragile and lacks robustness, showing heavy dependence on the split and order of input data. As shown in Fig~\ref{fig:boxplots}, in ten random trials, the worst trial shows a significant performance drop on new classes compared to the default setting. %selecting the most frequently occurring categories as base classes.
\end{itemize}
Therefore, in this work, we aim to understand the built-in objectness and achieve consistent improvements (especially on plasticity) across different task lengths and varying data input orders. This is crucial, as it is impractical to assume fixed task lengths and data sequences in real-world scenarios.
\begin{figure}
    \vspace{2pt}
    \centering
    \includegraphics[width=1\linewidth]{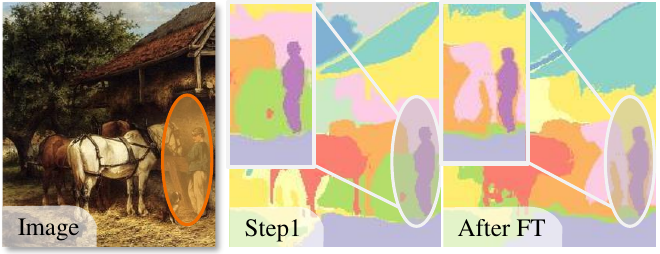}
    \caption{\textbf{Clustering results} from feature map. Pixel feature provides sufficient semantic priors (Person) even after finetuning.}
    \label{fig:clustering}
    \vspace{-10pt}
\end{figure}
The conclusion from a series of investigations is:
\begin{itemize}
    \item \textcolor{violet}{\ding{182}} \textbf{\textit{The built-in objectness emerges from the alignment between the query and the semantic priors within the image feature, mediated by the decoder.}} As shown in Fig~\ref{fig:clustering}, the clustering results indicate that image features contain sufficient semantic priors where pixels belonging to the same semantic are grouped together) even after finetune. Meanwhile, the query continuously aligns with specific regions of the feature map at each layer of the decoder as shown in Fig~\ref{fig:3} (right). In summary, the highly aggregated image feature provides a shortcut for queries to generate masks by simply aligning themselves to semantic priors in the image feature through the decoder. 
    \item \textcolor{violet}{\ding{183}} %\textbf{\textit{The built-in objectness diminishes across stages due to the failure of the query feature to capture specific regions of the feature map in the first layer.}} 
    \textbf{\textit{The built-in objectness diminishes over training stages due to the query's failure to align with the semantic priors of the feature map.}} 
    As shown in Fig~\ref{fig:3} (left), since semantic priors vary at different stages due to background semantic shift, causing the updated learnable query to gradually misalign with the pixel feature from old classes in previous stages, even after the decoder's post-alignment (observed in \textcolor{violet}{\ding{182}}). 
\end{itemize}

\noindent Inspired by \textcolor{violet}{\ding{182}} and \textcolor{violet}{\ding{183}}, to ensure objectness is preserved throughout the continual learning stages, we propose a \textbf{lazy Query Pre-Alignment (QPA)} method, where query features are selected from specific locations in the image feature map, rather than being learned from scratch, to ``\textit{perfectly}'' pre-align query feature with semantic priors. 
Specifically, based on the current stage's semantic classes, we select the most semantically significant locations in the image feature, preserving objectness at each stage. However, objectness is still lost across stages due to varying semantic classes in different stages.

To overcome cross-stage selection issues, 
a naive solution involves distillation on the feature map or query features between stages.
However, in turn, while it preserves old priors from previous stages, it re-introduces incorrect priors for current stages (where old priors label current semantics as background), leading to a loss of plasticity. 
Fortunately, thanks to our query pre-alignment method, we can easily maintain old classes by keeping queries corresponding to old class positions, while enabling the selection of remaining queries for new classes in the current stage. Thus, we propose a \textbf{Consistent Selection Loss (CSL)} to ensure that, for the same image, the most semantically significant locations selected in the previous stage are revisited in the current stage. 

With QPA and CSL, objectness in the query-based transformer is fully utilized to generate mask proposals. However, for class prediction, catastrophic forgetting may still occur. Previous methods typically rely on image replay to mitigate catastrophic forgetting. In contrast, thanks to our query pre-alignment, our query inherently contains category semantics. By storing the query feature, we can simulate specific semantics without requiring the actual image to contain the corresponding category. Therefore, we propose a novel \textbf{Virtual Query (VQ)} strategy to replay the virtual queries corresponding to previous classes in the decoder layer to avoid catastrophic forgetting. Compared to conventional image replay methods, our approach reduces storage requirements by $10$x, is independent of input data order, and preserves dataset privacy.

In summary, our contributions are multi-fold:
\begin{itemize}
    \item We provide a thorough analysis of the built-in objectness, revealing the reasons behind its emergence and demise.
    \item 
    By addressing the root cause, we can successfully leverage built-in objectness to mitigate catastrophic forgetting and background semantic shift through the introduction of three simple yet novel modules—QPA, CSL, and VQ.
    \item Our model, SimCIS, consistently and significantly outperforms state-of-the-art results on ADE20K in both continual panoptic and semantic segmentation. 
    \item We introduce new dataset splits to evaluate the model's robustness to input order in continual learning. SimCIS shows superior robustness over state-of-the-art methods, thanks to the effective utilization of built-in objectness. 
\end{itemize}

\section{Related Work}

\noindent \textbf{Continual Learning} is a longstanding field which possesses significant importance in addressing dynamic environments, enhancing model adaptability, and improving resource efficiency. The objective of continuous learning is to enable the model to efficiently acquire and adapt to new tasks and data, while retaining previously learned knowledge as it encounters additional information. The greatest challenge of continual learning is catastrophic forgetting~\cite{(cfcn)Robert1999catastrophic,(cfrp)robins1995catastrophic,(lla)thrun1998lifelong}. The early research are categorized into three primary types: those that rely on regularization constraints~\cite{(rwil)chaudhry2018riemannian,(lwm)dhar2019learning,(podnet)douillard2020podnet,(lwf)li2017learning,(gem)lopez2017gradient,(ellagem)chaudhry2018efficient}, those employing replay techniques~\cite{(learntoremember)ostapenko2019learning,(cldgr)shin2017continual,(cfrp)robins1995catastrophic}, and those based on dynamic structures~\cite{(Dytox)douillard2022dytox,(piggyback)mallya2018piggyback,(packnet)mallya2018packnet,(der)yan2021dynamically,(calicnn)singh2020calibrating,(slca)zhang2023slca}  . Regularization-based methods aim to reduce the interference of new tasks on old knowledge by constraining the learning process of the model, ensuring that the model parameters remain closely aligned with previously learned representations when updated due to task changes. Replay-based methods employ strategies to store, replay~\cite{(lucir)hou2019learning,(iCaRL)rebuffi2017icarl,(eteil)castro2018end,(lsil)wu2019large}, or generate~\cite{(cldgr)shin2017continual,(mrgan)wu2018memory,(learntoremember)ostapenko2019learning} samples from old tasks to mitigate catastrophic forgetting. Those methods based on dynamic structure~\cite{(packnet)mallya2018packnet,(piggyback)mallya2018piggyback,(prognw)rusu2016progressive} allocate distinct subsets of parameters to various subtasks by facilitating the expansion of their network architecture. 

\noindent \textbf{Universal Image Segmentation.}
Before MaskFormer proposed, traditional segmentation methods developed specialized architectures and models for each task to achieve top performance~\cite{(deeplab)chen2017deeplab,(siscnn)chen2014semantic,(maskrcnn)he2017mask,(sds)hariharan2014simultaneous,(carcnn)cai2018cascade,(segformer)xie2021segformer,(segmenter)strudel2021segmenter,(spgnet)cheng2019spgnet,(pspn)zhao2017pyramid,(ocnet)yuan2018ocnet,(ccnet)huang2019ccnet,(rethinking)chen2017rethinking}. MaskFormer~\cite{(maskformer)cheng2021per} is the first unified segmentation architecture to achieve state-of-the-art performance across three image segmentation tasks. Mask2Former~\cite{(mask2former)cheng2022masked} improves MaskFormer by adapting multi-scale features and introducing mask attention mechanism and achieve better performance. Follow its success in segmentation, we use Mask2Former as our baseline aims to extend its capability into the field of continual learning.

\noindent \textbf{Continual Segmentation} is the application of continual learning within the field of image segmentation. The challenge of continual segmentation tasks lies in the ability to identify new categories while generating high-quality masks for each category. This dual requirement underscores the complexity of maintaining accurate segmentation performance while adapting to an evolving set of class labels. Methods for continual segmentation are also categorized into three types as previously mentioned: regularization-based~\cite{(MiB)cermelli2020modeling,(PLOP)douillard2021plop,(SDR)michieli2021continual,(REMINDER)phan2022class,(incrementer)shang2023incrementer,(ewf)xiao2023ewf,(coinseg)zhang2023coinseg,(ILT)michieli2019incremental,(RC)zhang2022representation,(comformer)cermelli2023comformer}, replay-based~\cite{(ssul)cha2021ssul,(isbaseline)zhang2022mining,(cssmemory)zhu2023continual,(bacs)elaraby2024bacs,(bal)chen2024bal}, and dynamic structure-based~\cite{(CoMasTRe)gong2024continual,(decomposekd)baek2022decomposed,(awt)goswami2023attribution,(earlypayoff)xie2024early,(eclipse)kim2024eclipse}. Among these methods, those query-based architectures demonstrate notable performance. CoMFormer~\cite{(comformer)cermelli2023comformer} is the first query-based method in the field of continuous panoptic segmentation, employing distillation and pseudo label to combat catastrophic forgetting. CoMasTRe~\cite{(CoMasTRe)gong2024continual} is inspired by the methods of CoMFormer and, while maintaining the use of distillation loss, decouples mask and class predictions in continuous segmentation tasks. ECLIPSE~\cite{(eclipse)kim2024eclipse} adapts the strategy of VPT~\cite{(vpt)jia2022visual}, freezing the majority of model parameters and providing a set of trainable queries for fine-tuning across different tasks. BalConpas~\cite{(bal)chen2024bal} attempts to combat catastrophic forgetting by employing a method that combines feature-based distillation and a replay sample set, aiming to learn new classes without negatively impacting previously acquired knowledge.

\section{Preliminary}
\label{sec:pre}
\subsection{Problem Setting}
Following the same continual learning setting in~\cite{(comformer)cermelli2023comformer}
, we train our model over $T$ steps. At each step \( t \), the model \( \mathcal{M}^t \) has access only to a subset \( \mathcal{D}^t=\{\bm x^t, \bm y^t\} \) of the entire dataset \( \mathcal{D}^{1:T} \), where \(\bm x^t \in \mathbb{R}^{C \times H \times W} \) denotes the image at the current step and \(\bm y^t \) represents the corresponding annotations (where it can only contain annotations for classes \( \mathcal{C}^t \)). 
This setup, where each stage involves learning different classes, makes the model highly susceptible to catastrophic forgetting as it tends to lose previously acquired knowledge at each training step. Meanwhile, as the same image may appear across different learning steps with entirely different annotations, we also face the issue of so-called background shift~\cite{(MiB)cermelli2020modeling}. 
Given these challenges, our objective is to design a model \( \mathcal{M} \) such that, at any stage \( t \), the model \( \mathcal{M}^t \) not only effectively learns from \( \mathcal{D}^t \) but also preserve
the previous class knowledge from \( \mathcal{D}^{1:t-1} \).

\begin{figure*}[htp]
    \centering
    \includegraphics[width=0.95\linewidth]{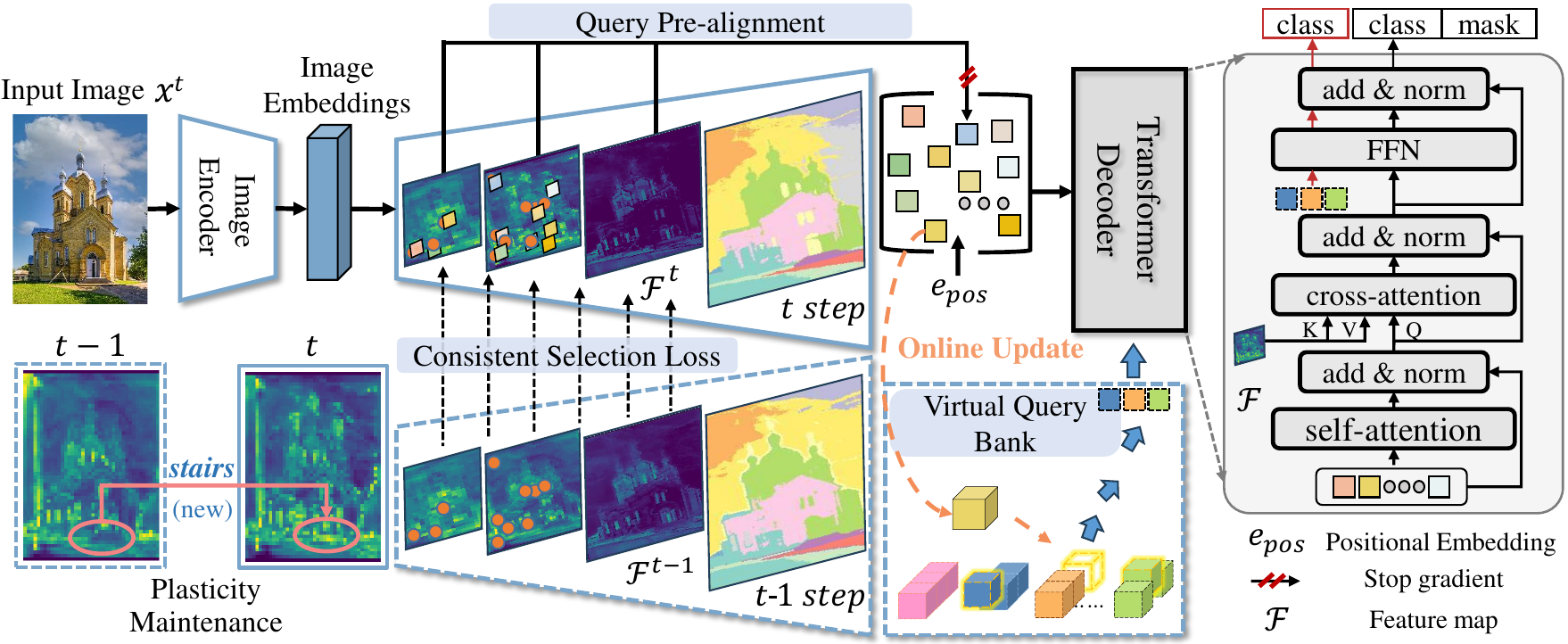}
    \caption{\textbf{The Overall Architecture of our SimCIS}: a lazy Query Pre-Alignment (Sec~\ref{sub:querypre}) with a Consistent Selection loss (Sec~\ref{sub:loss}) to ensure built-in objectness inner and across stages, and Virtual Query (Sec~\ref{sub:vq}) to avoid catastrophic forgetting in class prediction. }
    \label{fig:framework}
\end{figure*}

\subsection{Mask2Former}
\label{sub:mask2foremer}
We leverage Mask2former~\cite{(mask2former)cheng2022masked} as our meta-architecture for image segmentation.
Mask2Former is a transformer-based model, which predicts a set of binary masks instead of per-pixel classification, for universal segmentation tasks. It primarily consists of three components:
1) An image encoder as backbone \( {f}_{\text{backbone}} \) to extract image embeddings. 2) A pixel decoder \( {f}_{\text{pixel}} \) to embed image embeddings to multi-scale pixel features, which we denote as $F$:
\begin{equation}
       F = \{\mathcal{F}_{(l,h,w)} \,|\, \forall (l,h,w) \in \Omega\}, \,\,\mathcal{F} \in \mathbb{R}^{D \times H_l \times W_l},     
\end{equation}
where \( l \) denotes the multi-scale layer, \( D \) represents the hidden dimension, \( \mathcal{F}_{(l,h,w)} \) refers to the feature point at position \( (h, w) \) on the \( l \)-th layer and \( \Omega \) represents the spatial set of multi-scale features.
3) A transformer decoder \( f_{\text{decoder}} \) takes \( N \) learnable queries \( Q_N = \{ q_1, q_2, \dots, q_N \} \in \mathbb{R}^{N \times D} \) with positional encodings \( e_{pos} \in \mathbb{R}^{N \times D} \) to first conduct cross-attention and then self-attention with \( \mathcal{F} \) as follows:
\begin{equation}
    Q_N' =\text{FFN}(\text{SA}(\text{CA}(Q_N+e_{pos},F))),
\end{equation}
where \( \text{CA}(,) \) denotes the cross-attention, \( \text{SA}(\cdot) \) represents self-attention, and \( Q_N' \) denotes the updated query feature. The final prediction for each query is \( Z_N = \{ (c_i, m_i) \}_{i=1}^N \), where \( c_i \in \mathbb{R}^{C} \) and \( m_i \in \mathbb{R}^{H \times W} \) represent the predicted class and mask for \( q_i \), respectively. 

\section{Method}
\label{sec:method}
In this section, we introduce the overall architecture of our proposed SimCIS model for continual image segmentation. As shown in Fig~\ref{fig:framework}, SimCIS contains three modules: 1) Lazy Query Pre-alignment (Sec~\ref{sub:querypre}), 2) Consistent Selection Loss (Sec~\ref{sub:loss}) and Virtual Query (Sec~\ref{sub:vq}).

\subsection{Lazy Query Pre-alignment }
\label{sub:querypre}
To preserve the objectness across continual learning stages, we propose to pre-align the object query $Q_N$ with semantic priors in the pixel feature $\mathcal{F}_{(l,h,w)}$ by directly initializing query feature with the most semantically significant pixel feature.
To determine the semantic score of each pixel feature, we learn a prototype for each category and select pixel features as initial features by calculating the similarity between the pixel feature and each prototype.
Thus, we considered whether we could use explicit signals to guide the selection process of \( Q_N \) to enable the model to autonomously learn to choose the most discriminative points on the feature map.

Specifically, for each training step \( t \), we maintain a set of trainable prototypes \( \{ p^i \,|\, i \in C^t \} \), \( p^i \in \mathbb{R}^D \) for each class in $C^t$. By concatenating the prototypes of the past step, \( \mathcal{P}^{t-1} \), with those of the current classes, we obtain the current prototype set \( \mathcal{P}^t \) as follows,
\begin{equation}
 \mathcal{P}^t = \operatorname{concat}(\mathcal{P}^{t-1}, \{ p^i \,|\, i \in C^t \}).
\end{equation}
Then, for each feature point on \( F \), we compute its similarity with \( \mathcal{P}^t \) to select the best feature points. The selection process is as follows:
\begin{equation}
\mathcal{I}^{t} = \operatorname{topK}\left( \left\{ \max \, S(\mathcal{F}^{t}_{(l,h,w)}, \mathcal{P}^t) \mid \forall (l,h,w) \in \Omega \right\}, N \right),
\label{eq:qpa_select}
\end{equation}
\begin{equation}
Q_N = \mathcal{E}^{n=t}_{m=t}=\left\{ \mathcal{F}^{m=t}_i \mid i \in \mathcal{I}^{n=t} \right\}, 
\end{equation}
where \( \mathcal{I} = \{(l_i, h_i, w_i)\}_{i=0}^N \in \Omega \) represents the spatial positions of the selected feature points, \(\mathcal{E}^{n}_m\) represents 
the feature points from $\mathcal{F}^{m}$ selected by $\mathcal{I}^{n}$ and \( S(,) \) denotes the similarity calculation by dot product. The \(\operatorname{topK}(X,Y)\) function returns the indices of the \(Y\) largest values in \(X\) and \(N\) is the number of object query \( Q_N \) .
We select \( N \) feature points with the highest similarity with the prototype to initialize \( Q_N \).
To supervise our selection process, we use a classification loss during training and update \( \mathcal{P}^t \) through backpropagation \cite{rumelhart1986learning}.
% Compared to the initial naive approach, this method of query selection is more flexible. As shown in the figure, our selection process better leverages the advantages of multi-scale features. For objects of varying sizes, the location of the feature point can lie on different feature maps with varying receptive fields. 
Additionally, we apply stop gradient on \( Q_N \) to ensure that the information in \( F \) is not disrupted during training, keeping the objectness information stable across different stages.

\subsection{Consistent Selection Loss}
\label{sub:loss}
To ensure selection $\mathcal{I}$ is stable for the same image across stages, we propose a consistent selection loss.
% Due to our Lazy Query Pre-align strategy, which selects feature points \( \mathcal{E} \) on \( F \). We can easily obtain the spatial positions of \( \mathcal{E} \). In fact, our method relies on this spatial correspondence as a bridge to facilitate the transfer of information across different stages, thereby stabilizing object selection and avoiding the impact on plasticity caused by computing all feature points.
\textcolor{black}{Specifically, when training our model \( \mathcal{M}^t \) at current stage, \textcolor{black}{we can easily obtain feature points} \(  \mathcal{E}^{t-1}_{t}=\{\mathcal{F}^{t}_i\mid i \in \mathcal{I}^{t-1}\}\) }.
% \todo{by obtaining the spatial position indices \( \mathcal{I}^{t-1} \) selected by \( M^{t-1} \). We can easily obtain the corresponding feature points at the current stage \( \mathcal{E}^{t-1}_{t} \) as:}
Then, to maintain consistency in object selection across different steps, we calculate the similarity between selected feature points with \( \mathcal{P}^{t-1} \), after that, we use the Kullback-Leibler (KL) divergence loss \cite{hinton2015distilling} to compute the loss:
\begin{equation}
L_{csl}=\frac{1}{|\mathcal{I}^{t-1}|} \sum_{i=1}^{|\mathcal{I}^{t-1}|} S(\mathcal{E}^{t-1}_{t-1}, \mathcal{P}^{t-1}) \log \frac{S(\mathcal{E}^{t-1}_{t-1}, \mathcal{P}^{t-1})}{S(\mathcal{E}^{t-1}_{t}, \mathcal{P}^{t-1})}.
\label{eq:csl}
\end{equation}
In this way, we successfully maintain the most semantically significant locations from the previous stage, ensuring that the selection of \( Q_N \) remains stable across stages.

\subsection{Virtual Query}
\label{sub:vq}
To overcome catastrophic forgetting in class prediction, we propose the virtual query to bypass the limitations of previous methods that rely on data order.
Virtual Query replays the previous query feature in the decoder layer to simulate semantics.
Specifically, our innovative virtual query strategy can be divided into three steps: \textcolor{black}{Firstly, we use the results of bipartite matching to select object queries and build our VQ bank. Then we analyze the pseudo-distribution to focus on rare categories in the current stage. Finally, we sample VQs in the new stage according to the pseudo-distribution and concatenate them into the object query \( Q_N \) for input into the decoder.
}

\noindent\textbf{(1) Query Storage. }During training, we maintain a queue of length \( h \) for each class, forming our virtual query bank
\begin{equation}
    \mathcal{B}_{\text{vq}} = \{b_1^h, b_2^h, \dots, b_{|c^{1:T}|}^h\},
\end{equation}
 \textcolor{black}{where \(b_i^h\) represents a queue of length \(h\) for class \(i\)} where \(b_i^h\) is the queue for class \(i\). Queries matched through bipartite matching \cite{(detr)carion2020end} from the decoder's final layer output, \( Z_N \) \textcolor{black}{(defined in Sec~\ref{sub:mask2foremer})}, are stored in the appropriate class queues based on their bipartite matching results with ground truth \(\bm y\).
\begin{equation}
\left\{
\begin{array}{l}
    \mathcal{I}_b = \operatorname{Bipartite}(Z_N, \bm y), \\[8pt]
    \mathcal{B}_{\text{vq}} \leftarrow \underset{\forall i = (i_q, i_y) \in \mathcal{I}_b}{\operatorname{Enqueue}}(Q_N(i_q), b_{\hat{y}^{(i_y)}}),
\end{array}
\right.
\label{eq:update_vq}
\end{equation}
where \( N \) denotes the number of queries. The set \( \mathcal{I}_b \) consists of tuples, where each tuple \( i = (i_q, i_y) \) represents the correspondence between query and ground truth. Here, \( i_q \) denotes the query index, and \( i_y \) denotes the ground truth index. \textcolor{black}{\( \hat{y}^i \) represents the class label of the \(i^{th}\) ground truth. }

\noindent\textbf{(2) Pseudo-Distribution Statistics. }In each continual learning step, the category distribution of images changes at each stage. To ensure the decoder retains the category information for all old classes,  \textcolor{black}{we use the pre-trained \textcolor{black}{last-stage model} \(\mathcal{M}^{t-1}\)'s outputs on \textcolor{black}{current stage's dataset} \(D^t\) to simulate the distribution of real classes which helps mitigate the forgetting of rare classes in the current stage. We use this pseudo-distribution statistics by calculating}

\begin{equation} 
\omega = \left\{ \left( (\sum_{i=1}^m \sigma_i) / \sigma_j \right)^{\frac{1}{2}} \right\}_{j=1}^m ,
\label{eq:pds}
\end{equation}
\textcolor{black}{where \(\sigma_i\) is the pseudo number of class \(i\) in the current stage and \( m = |c^{1:t-1}| \) represents the number of categories from the previous stages. }

\noindent\textbf{(3) VQ Utilization.} Based on the pseudo-distribution statistics, in each iteration, we sample \( j \) virtual queries \( Q_j = \{vq_1, \dots, vq_j\} \) for each batch based on \(\omega\). These queries are then concatenated with  \( Q_N \) as
\begin{equation}
    Q_{N+j} =\{q_1,\cdots,q_N,vq_1,\cdots,vq_j\},
\end{equation}
and fed into the decoder.
As shown in Fig~\ref{fig:framework}, within the decoder, we design a skip attention strategy for the VQs. Specifically, since the objects represented by the VQs do not appear in the image, to prevent the VQs from influencing \( Q_N \) during the self-attention and cross-attention processes, we allow the VQs to bypass the attention layers and directly affect the FFN layers as follows:
\begin{equation}
    Q_{N+j}' =\text{FFN}(\operatorname{concat}[\text{CA}(\text{SA}(Q_N+e_{pos},F)),Q_j]).
    \label{eq:skip}
\end{equation}
Finally, the virtual query only computes \(L_{\text{class}}\) to address the model's category forgetting.
\section{Experiments}
\label{exp}
\subsection{Experimental Setup}
\begin{table*}[t]
      \centering
      \begin{adjustbox}{max width=0.95\linewidth}
      \begin{tabular}{c|cccc|cccc|cccc}
        \toprule
        \multirow{2}{*}{Method}  & \multicolumn{4}{c|}{\textbf{100-5} (11 tasks)} & \multicolumn{4}{c|}{\textbf{100-10} (6 tasks)} & \multicolumn{4}{c}{\textbf{100-50} (2 tasks)} \\
         & \textit{1-100} & \textit{101-150} & \textit{all} & \textit{avg} & \textit{1-100} & \textit{101-150} & \textit{all} & \textit{avg} & \textit{1-100} & \textit{101-150} & \textit{all} & \textit{avg}\\
        \midrule
        FT & 0.0 & 2.2 & 0.7 & 4.7 & 0.0 & 4.8 & 1.6 & 8.9  & 0.0 & 32.4 & 10.8 & 26.8 \\
        MiB~\cite{(MiB)cermelli2020modeling} & 2.3 & 0.0 & 1.5& 13.4 & 6.8 & 0.2 & 4.6 & 19.1 & 23.3 & 14.9 & 20.5& 31.7 \\
        PLOP~\cite{(PLOP)douillard2021plop}  & 31.1 & 11.9 & 24.7 & 31.3 & 37.7 & 23.3 & 32.9 & 37.8 & 42.4 & 23.7 & 36.2 & 39.5\\
        SSUL~\cite{(ssul)cha2021ssul}  & 30.2 & 7.9 & 22.8 & 27.9 & 31.6 & 11.9 & 25.0 & 30.3 & 35.9 & 18.1 & 30.0 & 33.8\\
        CoMFormer \cite{(comformer)cermelli2023comformer} & 34.4 & 15.9 & 28.2 & 34.0 & 36.0 & 17.1 & 29.7 & 35.3 & 41.1 & 27.7 & 36.7 & 38.8\\
       BalConpas \cite{(bal)chen2024bal} & 36.1 & \underline{20.3} & 30.8 & 35.8 & 40.7 & \underline{22.8} & \underline{34.7} & 38.8 & \underline{42.8} & \underline{25.7} & \underline{37.1} & 40.0 \\
       ECLIPSE \cite{(eclipse)kim2024eclipse} & \underline{41.1} & 16.6 & \underline{32.9} & - & \underline{41.4} & 18.8 & 33.9 & - & 41.7 & 23.5 & 35.6  & - \\
        % \midrule
        Our SimCIS & \tablefirst\textbf{42.1} & \tablefirst\textbf{21.9} & \tablefirst\textbf{35.4} & \tablefirst\textbf{38.7} & \tablefirst\textbf{42.2} & \tablefirst\textbf{30.1} & \tablefirst\textbf{38.1} & \tablefirst\textbf{40.5} &  \tablefirst\textbf{44.7} & \tablefirst\textbf{30.8} & \tablefirst\textbf{40.0} & \tablefirst\textbf{42.7} \\
        \midrule
        joint & 43.6 & 34.2 & 40.4 & - & 43.6 & 34.2 & 40.4 & - & 43.6 & 34.2 & 40.4 & - \\ % our joint
        \bottomrule
      \end{tabular}
      \end{adjustbox}
  \caption{
    \textbf{Continual Panoptic Segmentation} results on ADE20K dataset in PQ. All methods use the same network of Mask2Former~\cite{(mask2former)cheng2022masked} with ResNet-50~\cite{(resnet)he2016deep} backbone. \textit{joint} means an oracle setting training all classes offline at once.
  }
  \label{tab:1}
\end{table*}
\begin{table*}[t]
      \centering
      \begin{adjustbox}{max width=0.8\linewidth}
      \begin{tabular}{c|ccc|ccc|ccc}
        \toprule
        \multirow{2}{*}{Method}& \multicolumn{3}{c|}{\textbf{50-10} (11 tasks)} & \multicolumn{3}{c|}{\textbf{50-20} (6 tasks)} & \multicolumn{3}{c}{\textbf{50-50} (3 tasks)} \\
         & \textit{1-50}~~ & ~~\textit{51-150} & \textit{all} & \textit{1-50}~~ & ~~\textit{51-150} & \textit{all} & \textit{1-50}~~ & ~~\textit{51-150} & \textit{all} \\
        \midrule
        FT  & 0.0  & 1.7  & 1.1 & 0.0  & 4.4  & 2.9 & 0.0 & 12.0 & 8.1     \\
        MiB~\cite{(MiB)cermelli2020modeling} & 34.9  & 7.7  & 16.8 & 38.8  & 10.9  & 20.2 & 42.4 & 15.5 & 24.4  \\
        PLOP~\cite{(PLOP)douillard2021plop} & 39.9 & 15.0  & 23.3 & 43.9  & 16.2  & 25.4  & 45.8 & 18.7 & 27.7   \\
        CoMFormer~\cite{(comformer)cermelli2023comformer}  & 38.5 & 15.6 & 23.2 & 42.7 & 17.2 & 25.7  & 45.0 & 19.3 & 27.9  \\
        ECLIPSE \cite{(eclipse)kim2024eclipse} & 45.9 &17.3 &26.8 & 46.4 & 19.6 &28.6  &46.0 & 20.7 & 29.2     \\
        BalConpas~\cite{(bal)chen2024bal} & 44.6 & 24.8 & 31.4  & 49.2 & 28.2 & 35.2  & 51.2 & 26.5 & 34.7  \\
       Our SimCIS & \tablefirst\textbf{48.8} & \tablefirst\textbf{30.0} & \tablefirst\textbf{36.3}  & \tablefirst\textbf{51.6} & \tablefirst\textbf{31.9} & \tablefirst\textbf{38.5}  & \tablefirst\textbf{52.1} & \tablefirst\textbf{30.7} & \tablefirst\textbf{37.9}  \\
        \midrule
        joint & 51.1 & 35.1 & 40.4 &51.1 & 35.1 & 40.4   & 51.1 & 35.1 & 40.4   \\
        \bottomrule
      \end{tabular}
      \end{adjustbox}
  \caption{
    \textbf{Continual Panoptic Segmentation} results on ADE20K dataset in PQ. All methods use Mask2Former~\cite{(mask2former)cheng2022masked} with ResNet-50~\cite{(resnet)he2016deep}.
  }
  \label{tab:2}
  \vspace{-5pt}
\end{table*}
\textbf{Dataset and Evaluation Metric.} Following previous works~\cite{(comformer)cermelli2023comformer,(eclipse)kim2024eclipse,(bal)chen2024bal}, we compare our SimCIS with other approaches using the ADE20K dataset~\cite{(ade20k)zhou2017scene} to evaluate its effectiveness.
% The ADE20K dataset contains $20,210$ training images and $2,000$ validation images, with each image averaging $19.5$ instances and $10.5$ classes. 
The images in the dataset include annotations for $150$ classes, which are ranked by their total pixel ratios in the whole dataset. Among these $150$ classes, $50$ amorphous background classes are labeled as ``stuff'' classes, while $100$ discrete object classes are labeled as ``thing'' classes. 
Following\cite{(comformer)cermelli2023comformer}, we use Panoptic Quality (PQ) as the performance metric for continual panoptic segmentation and mean Inter-over-Union (mIoU) for continual semantic segmentation. After incremental learning steps, 
we report results for base classes ($\mathcal{C}^1$), new classes ($\mathcal{C}^{2:T}$), 
all classes ($\mathcal{C}^{1:T}$), 
and an average of all visible classes at each step (avg), respectively.

\noindent\textbf{Continual Learning Protocol.} Following existing continual segmentation methods~\cite{(MiB)cermelli2020modeling,(PLOP)douillard2021plop,(ssul)cha2021ssul,(comformer)cermelli2023comformer,(bal)chen2024bal,(eclipse)kim2024eclipse,(CoMasTRe)gong2024continual}, we evaluate our method on different continual learning settings. In particular, our incremental learning tasks are represented in the form of \(A\)-\(B\), where \(A\) denotes the number of base classes partitioned from the dataset, and \(B\) denotes the number of new classes. 
For both continual panoptic (CPS) and semantic segmentation (CSS), we conduct tasks of $100$ - $5$, $100$ - $10$, and $100$ - $50$. Additionally, we conduct tasks of $50$ - $10$, $50$ - $20$, and $50$ - $50$ for panoptic segmentation.

\noindent\textbf{Implementation Details.} We adapt an pre-trained ResNet-50~\cite{(resnet)he2016deep} backbone for CPS and an pre-trained ResNet-101 for CSS. Following previous work~\cite{(bal)chen2024bal}, the input image resolution for the CPS tasks is set to \(640 \times 640\), while for the CSS tasks, it is set to \(512 \times 512\). For the number of virtual queries \(N\), it be set up to 80.
For more detailes, please refer to the Appendix. 
\subsection{Quantitative Results}
Tab~\ref{tab:1}, Tab~\ref{tab:2} and Tab~\ref{tab:3} present the performance of SimCIS and other approaches on the continual panoptic segmentation and semantic segmentation benchmark. In these tables, ``FT'' refers to fine-tuning the base model without employing continual learning methods, while ``joint'' indicates training the base model using all available data. They represent the lower and upper-performance bounds for continual learning methods, respectively.
\begin{table*}[t]
\centering
  \begin{adjustbox}{max width=0.95\linewidth}
  \begin{tabular}{c|cccc|cccc|cccc}
    \toprule
    \multirow{2}*{Model} & \multicolumn{4}{c}{\textbf{100-5} (11 tasks)} & \multicolumn{4}{c}{\textbf{100-10} (6 tasks)} & \multicolumn{4}{c}{\textbf{100-50} (2 tasks)} \\
    & \textit{1-100}~~ & ~~\textit{101-150} & \textit{all} & \textit{avg} & \textit{1-100}~~ & ~~\textit{101-150} & \textit{all} & \textit{avg} & \textit{1-100}~~ & ~~\textit{101-150} & \textit{all} & \textit{avg} \\
    \midrule
    FT & 0.0 & 0.3 & 0.1 & 5.6 & 0.0 & 0.1 & 0.0 & 9.1 & 0.0 & 3.2 & 1.1 & 26.3 \\
    MiB \cite{(MiB)cermelli2020modeling} & 36.0 & 5.7 & 26.0 & - & 31.8 & 14.1 & 25.9 & - & 37.9 & 27.9 & 34.6 & - \\
    PLOP \cite{(PLOP)douillard2021plop} & 39.1 & 7.8 & 28.8 & 35.3 & 40.5 & 14.1 & 31.6 & 36.6 & 41.9 & 14.9 & 32.9 & 37.4 \\
    SSUL \cite{(ssul)cha2021ssul} & 42.9 & 17.8 & 34.6 & - & 42.9 & 17.7 & 34.5 & - & 42.8 & 17.5 & 34.4 & - \\
    EWF \cite{(ewf)xiao2023ewf} & 41.4 & 13.4 & 32.1 & - & 41.5 & 16.3 & 33.2 & - & 41.2 & 21.3 & 34.6 & - \textbf{}\\
    CoMFormer \cite{(comformer)cermelli2023comformer} & 39.5 & 13.6 & 30.9 & 36.5 & 40.6 & 15.6 & 32.3 & 37.4 & 39.5 & 26.2 & 38.4 & 41.2 \\
    ECLIPSE \cite{(eclipse)kim2024eclipse} & 43.3 & 16.3 & 34.2 & - & 43.4 & 17.4 & 34.6 & - & 45.0 & 21.7 & 37.1 & - \\
    BalConpas \cite{(bal)chen2024bal} &42.1& 17.2 &33.8&41.3&47.3& 24.2 &38.6 &43.6 &  49.9 & 30.1 & 43.3 & 47.4 \\
    CoMasTRe \cite{(CoMasTRe)gong2024continual} &40.8& 15.8 &32.6&38.6&42.3& 18.4 &34.4 &38.4 &  45.7 & 26.0 & 39.2 & 41.6 \\
    Our SimCIS & \tablefirst\textbf{46.7} & \tablefirst\textbf{22.8} & \tablefirst\textbf{38.7} & \tablefirst\textbf{47.4} & \tablefirst\textbf{49.7} & \tablefirst\textbf{27.4} & \tablefirst\textbf{42.3} & \tablefirst\textbf{49.2} & \tablefirst\textbf{54.9} & \tablefirst\textbf{36.0} & \tablefirst\textbf{48.6} & \tablefirst\textbf{52.0}
        \\
    \midrule
    
    Joint & 57.1 & 39.1 & 51.2 & - & 57.1 & 39.1 & 51.2 & - & 57.1 & 39.1 & 51.2 & - \\
    \bottomrule
  \end{tabular}
   \end{adjustbox}
  \caption{\textbf{Continual Semantic Segmentation} results on the ADE20K dataset, measured by mIoU.}
  \label{tab:3}
\end{table*}
  
\noindent\textbf{Continual Panoptic Segmentation.} 
Tab~\ref{tab:1} and Tab~\ref{tab:2} present the performance of SimCIS and other approaches under different continual panoptic segmentation settings. \textbf{(1)} Compared to regularization-based methods MiB~\cite{(MiB)cermelli2020modeling}, PLOP~\cite{(PLOP)douillard2021plop}, and CoMFormer~\cite{(comformer)cermelli2023comformer}, SimCIS achieves superior results on both new and base classes. Notably, compared to CoMFormer, the best-performing among them, SimCIS improves PQ by $+6.0$\% on new classes and $+7.7$\% on base classes in the $100$ - $5$ task, maintaining a consistent lead in the $100$ - $10$ and $100$ - $50$ tasks. Especially in the $100$ - $10$ task, it surpasses CoMFormer by $+6.2$\% PQ on base and $+13.0$\% PQ on new classes. When using $50$ base classes, SimCIS significantly outperforms these methods, demonstrating its superiority. \textbf{(2)} Compared with the method also using built-in objectness, SimCIS achieves better performance on new classes without freezing the model parameters. In the $ 100$ - $5$ , $ 100$ - $10$ , and $ 100$ - $50$  tasks, SimCIS outperforms ECLIPSE~\cite{(eclipse)kim2024eclipse} by $+5.3$\%  PQ, $ +11.3$\%  PQ, and $ +7.6$\%  PQ, respectively. In the tasks with $ 50$ classes as base classes, SimCIS outperforms ECLIPSE~\cite{(eclipse)kim2024eclipse} by over $ +10$\%  PQ on new classes, demonstrating the stability of our approach. \textbf{(3)} BalConpas~\cite{(bal)chen2024bal} is a continual learning method based on the Mask2Former~\cite{(mask2former)cheng2022masked} architecture. In the $ 100$ - $10$  and $ 100$ - $50$  tasks, SimCIS outperforms BalConpas~\cite{(bal)chen2024bal} by more than $ +5.0$\%  PQ on new classes. In the longer step sequence of the $ 100$ - $5$ task, SimCIS surpasses BalConpas~\cite{(bal)chen2024bal} by $ +6.0$\%  PQ on base classes. In the $ 50$ - $20$  and $ 50$ - $50$  tasks, SimCIS maintains strong performance, averaging $ +4$\%  PQ higher than BalConpas~\cite{(bal)chen2024bal} on new classes. In the longer step sequence of the $ 50$ - $10$ task, SimCIS exceeds BalConpas~\cite{(bal)chen2024bal} by $ +4.2$\%  PQ on base classes. It is noteworthy that in the $ 100$ - $50$  task, SimCIS almost matches the performance of the ``joint'', with base classes performance even exceeding that of the ``joint''.

\noindent\textbf{Continual Semantic Segmentation.} 
As shown in Tab~\ref{tab:3}, we further compare SimCIS with state-of-the-art works in continual semantic segmentation. \textbf{(1)} Across three tasks, SimCIS surpasses prior approaches by at least $+4$\% mIoU on base classes. For new classes, it outperforms SSUL~\cite{(ssul)cha2021ssul} by $+5.0$\% and $+9.7$\% mIoU in the $100$ - $5$ and $100$ - $10$ tasks, respectively. In the $100$ - $50$ task, SimCIS surpasses MiB~\cite{(MiB)cermelli2020modeling}, which achieves $27.9$\% mIoU, by $+8.1$\% mIoU. \textbf{(2)} Among Mask2Former~\cite{(mask2former)cheng2022masked}-based methods, SimCIS also achieves the best results. In the $100$ - $5$ task, it outperforms ECLIPSE~\cite{(eclipse)kim2024eclipse} on base classes by $+3.4$\% mIoU and BalConpas~\cite{(bal)chen2024bal} on new classes by $+5.6$\% mIoU. In the $100$ - $10$ task, SimCIS achieves the performance of new classes exceeding all other architectures by at least $+3.0$\% mIoU while maintaining high performance on base classes. 
\begin{table}[t]
  \centering
      \begin{adjustbox}{max width=\linewidth}
      \begin{tabular}{c|c|c|c|ccc|ccc}
        \toprule
        \multirow{2}{*}{Psd} & \multirow{2}{*}{QPA} & \multirow{2}{*}{CSL} & \multirow{2}{*}{VQ} & \multicolumn{3}{c}{\textbf{Panoptic 100-5} (11 tasks)} & \multicolumn{3}{c}{\textbf{Semantic 100-5} (11 tasks)}\\
         &  &   &  & \textit{1-100} & \textit{101-150} & \textit{all} & \textit{1-100} & \textit{101-150} & \textit{all}\\
        \midrule
        % &  &  & & 0.2 & 3.9 & 1.3 & 0.0 & 0.3 & 0.1\\
        \checkmark & &  &  & 31.6& 21.3 & 28.2 & 15.6 & 8.5 & 13.2\\
        \checkmark & \checkmark & &  & 30.7 & 22.3 & 27.9 &  37.4 & 16.7 & 30.5\\
        % \checkmark & \checkmark & \checkmark & & 38.9 & 16.0 & 31.2 & 43.2 & 17.0 & 34.5 \\
        \checkmark & \checkmark & \checkmark & & 35.7 & 24.0 & 31.8 & 43.2 & 17.0 & 34.5 \\
        \checkmark & \checkmark &  & \checkmark & 35.1 & 23.3 & 31.2 & 42.5 & 19.5 & 34.8\\
         \checkmark & \checkmark & \checkmark & \checkmark & 42.1 & 21.9 & 35.4 & 46.7 & 22.8 & 38.7\\
        \bottomrule
      \end{tabular}
      \end{adjustbox}
      \caption{
        \textbf{Ablation Study on Proposed Components.} Psd: pseudo label, QPA: lazy query pre-alignment, CSL: consistent selection loss, and VQ: virtual query.
      }
      \label{tab:abla}
          \vspace{-10pt}
\end{table}

\begin{figure*}[t]
    \centering
    \includegraphics[width=0.9\linewidth]{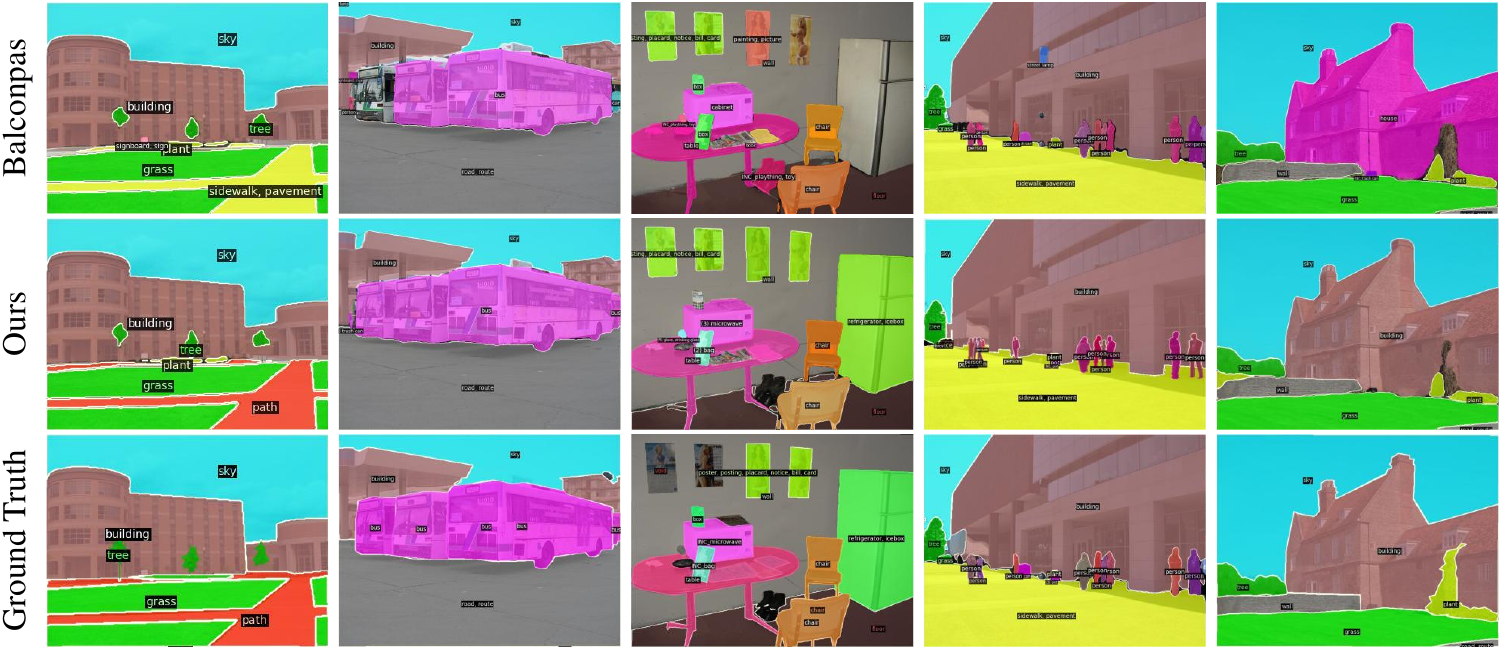}
    \caption{
        \textbf{Qualitative comparisons} between SimCIS and BalConpas~\cite{(bal)chen2024bal} on the ADE20K 100-5 continual panoptic segmentation scenario. Our SimCIS demonstrates significant results, highlighting the effectiveness of our strategies.}
    \label{fig:1}
       \vspace{-10pt}
\end{figure*}

\subsection{Qualitative Comparison.}
\noindent\textbf{Comparison with Previous SOTAs.} We compare SimCIS with BalConpas~\cite{(bal)chen2024bal} in the $100$ - $5$ continual panoptic segmentation task of the ADE20K dataset, and the visual results are illustrated in Fig~\ref{fig:1}.  
In the first, second, and fifth examples, BalConpas~\cite{(bal)chen2024bal} encounters forgetting on base classes such as path, bus, and building. Additionally, in the third example, BalConpas incorrectly classifies the microwave and bag as cabinet and box, respectively. 
% These phenomena indicate that the classification capability of BalConpas~\cite{(bal)chen2024bal} is significantly affected during continual learning steps. 
Benefiting from the VQ, our SimCIS has a significant advantage in preserving class information, allowing it to perform well in these examples. Furthermore, BalConpas~\cite{(bal)chen2024bal} fails to provide segmentation masks for the bus and refrigerator instances in the second and third examples. In contrast, our proposed the keep built-in objectness strategy effectively preserves object information within the encoder, enabling SimCIS to accurately segment object instances. 

\begin{figure}[t]
    \centering
    \includegraphics[width=1\linewidth]{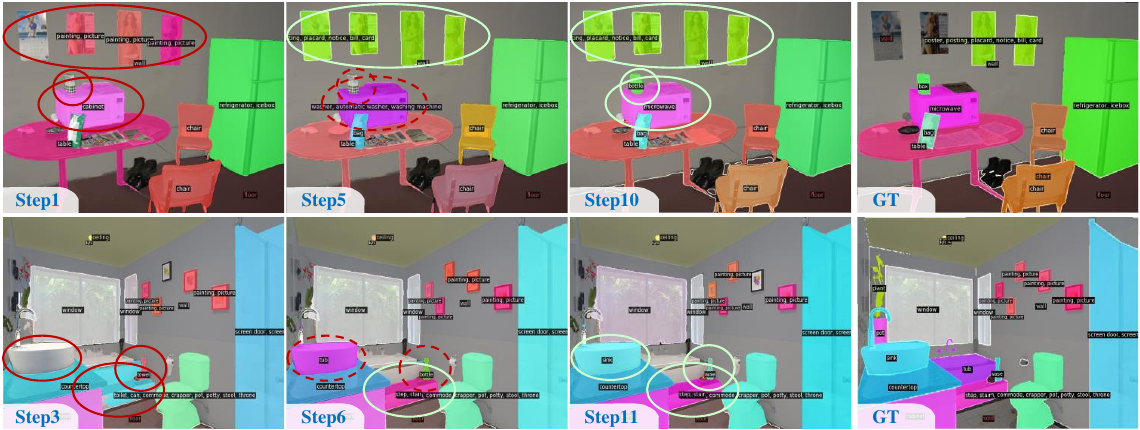}
    \caption{\textbf{Qualitative examples in continual learning.} 
    }
    \label{fig:enter-label}
    \vspace{-5pt}
\end{figure}

\noindent\textbf{Comparison in Different Steps.} To further illustrate the effectiveness of our method, we select certain visual examples from the continual learning steps of the $100$ - $5$ task. In the two examples shown in Fig~\ref{fig:enter-label}, our method is able to correct errors during the continual learning steps, such as the microwave and bag in the first image, as well as the sink, vase, and stair in the second image. SimCIS refines itself during the continual learning process, ultimately achieving accurate classification and segmentation of object instances based on our proposed flexible VQ.

\subsection{Ablation Study}
In this section, we report the results of the ablation experiments to validate the effectiveness of each component and configuration in our SimCIS. We select the $100$ - $5$ task in CPS and CSS to report the performance of SimCIS.

\noindent\textbf{Main Components.} 
As shown in Tab~\ref{tab:abla}, each component contributes to the overall performance.
We take Mask2Former~\cite{(mask2former)cheng2022masked} with pseudo label as our baseline performance. The second row of the table shows the performance of QPA with an increase of $+18.2$\% mIoU on base classes and an increase of $+8.2$\% mIoU on new classes.
With the help of CSL (the third row), the CSL strategy achieves increases of $+8.2$\% PQ and $+5.8$\% mIoU for base classes, respectively.
\begin{table}[t]
  \centering
          \begin{adjustbox}{max width=\linewidth}
          \begin{tabular}{c|c|c|c|c|c}
            \toprule
            Reply & Num & Disk & \multicolumn{2}{c}{\textbf{100-5} (11 tasks)} \\
              Type & Samples & Memory& \textit{base} &\textit{all} \\
            \midrule
            \multirow{5}{*}{Image}
            & 0 \footnotesize{~~~~~(*20)}& 0.0MB & 35.7 & 31.8 \\
             & 75 \footnotesize{~~~(*20)}& 3.4MB & 38.9 & 33.4 \\
             & 150 \footnotesize{~(*20)}& 6.1MB & 38.9  & 34.0 \\
             & 300 \footnotesize{~(*20)}& 11.8MB & 38.5 & 33.7 \\
             & 600 \footnotesize{~(*20)}& 21.9MB & 39.2 & 34.3 \\
            \midrule
            \multirow{5}{*}{Virtual Query} & 0 \footnotesize{~~~~(*150)}  & ~~0.0MB  &  35.7 & 31.8 \\
                                & 20 \footnotesize{~~(*150)}  & ~~1.5MB  &  40.6  & 34.6 \\
                                  & 40 \footnotesize{~~(*150)} & ~~3.0MB & 40.4  & 34.1 \\
                                  & 80 \footnotesize{~~(*150)} & ~~5.9MB& \tablefirst\textbf{42.1} & \tablefirst\textbf{35.4} \\
                                  & 160 \footnotesize{(*150)}  & ~~12.0MB& 40.9 & 34.2 \\
            
            \bottomrule
          \end{tabular}
          \end{adjustbox}
  \caption{
        \textbf{Effect of Replay Type} and \textbf{Storage Requirements}.
      }
      \label{tab:vq}
       \vspace{-10pt}
\end{table}

\noindent\textbf{Effectiveness of VQ.} 
As shown in Tab~\ref{tab:vq}, compared to the conventional image replay method, our VQ strategy demonstrates significant improvements in both storage efficiency and performance. Firstly, when using $300$ samples for the image replay and $80$ samples for VQ, we achieve a $+1.4$\% increase in PQ  across all classes while using almost the same disk memory.
When comparing the optimal cases for both storage methods, our VQ strategy outperforms the conventional image replay method by $+1.1$\% PQ, while utilizing only $27$\% of the storage space.

\noindent\textbf{Robust to Input Data Order.}
\begin{table}[t]
      \centering
      \begin{adjustbox}{max width=0.7\linewidth}
      \begin{tabular}{c|ccc}
        \toprule
        \multirow{2}{*}{Method}  & \multicolumn{3}{c}{\textbf{100-10} (6 tasks)} \\
          & \textit{1-100} & \textit{101-150} & \textit{all}\\
        \midrule
       BalConpas \cite{(bal)chen2024bal} & \underline{38.9(39.4)} & \textbf{27.8(26.8)} & \underline{35.2} \\
       ECLIPSE \cite{(eclipse)kim2024eclipse} & 32.7(32.1) & 22.3(23.8)& 29.3  \\
        % \midrule
        Ours  &  \textbf{40.3(40.2)} & \underline{25.4(25.7)} & \textbf{35.3} \\
        \midrule
        Joint  & (43.6) & (34.2) & (40.4) \\ % our joint
        \bottomrule
      \end{tabular}
      \end{adjustbox}
  \caption{
    \textbf{Continual Panoptic Segmentation} with random order. We also report the performance evaluated in the original class order in \((\cdot)\). For detailed experiments, please refer to the Appendix.
  }
  \label{tab:order}
     \vspace{-10pt}
\end{table}
As shown in Tab \ref{tab:order}, our model has great robustness in random data order. We have a $+0.1$\% PQ increase compared to BalConpas and a $+6.0$\% PQ increase against ECLIPSE across all classes. 

\section{Conclusion}
In this work, we present a novel class-incremental image segmentation (CIS) method called SimCIS, which addresses the challenges of catastrophic forgetting and background shift.
We first explore the emergence and diminishing of built-in objectness in query-based transformers and then propose two novel modules: lazy query pre-alignment and consistent selection loss, to ensure both intra-stage and cross-stage built-in objectness. Additionally, we introduce virtual queries to mitigate catastrophic forgetting in class prediction. 
Comparisons with previous state-of-the-art CIS methods and our ablation study demonstrate the superiority of each individual component in our model, highlighting its effectiveness in overcoming the challenges of incremental learning. \textbf{Acknowledgment:} 
This work was supported by the National Natural Science Foundation of China (No.62206174).

\clearpage
{
    \small
    \bibliographystyle{ieeenat_fullname}
    \bibliography{main}
}
\maketitlesupplementary

In this supplementary material, we provide additional information regarding:
\begin{itemize}
\item Overall Workflow of our SimCIS with Pseudocode (In Sec.~\ref{sec:suppl_pseudo}).
\item More Dataset and Implementation Details (In Sec.~\ref{sec:suppl_imp}).
\item Comprehensive Experiments of Random Class Rrder (In Sec.~\ref{sec:suppl_order}).
\item More Ablation Studies on the Stop-Gradient Strategy. (In Sec.~\ref{sec:suppl_stopg}).
\item More Visualization Results of the Continual Semantic Segmentation task (In Sec.~\ref{sec:suppl:css}).
\item More Visualization Results of Objectness Information (In Sec.~\ref{sec:suppl_kmeans}).
\item Discussion, Limitation and Future Work (In Sec.~\ref{sec:suppl_discussion}).
\end{itemize}

\section{Pseudocode for our SimCIS}
\label{sec:suppl_pseudo}
In this section, we present the overall workflow of our method in the pseudo-code Algo.~\ref{AL:1}. At the beginning, we define some modules, functions, and variables. For the current stage $t$ and the previous stage $t-1$, we define the backbone modules $f_{\text{backbone}}^t$ and $f_{\text{backbone}}^{t-1}$, the pixel decoder modules $f_{\text{pixel}}^t$ and $f_{\text{pixel}}^{t-1}$, the prototypes $\mathcal{P}^t$ and $\mathcal{P}^{t-1}$ respectively. 
For clarity and readability of the pseudocode, some formulas introduced in the main text are encapsulated as functions. These include the select feature points function $\Phi$ (Eq. $4$), the consistent selection loss function $l_{\text{csl}}$ (Eq. $6$), the calculate sample weights function $g$ (Eq. $9$), the virtual query bank $\mathcal{B}_{vq}$ update function $\mathcal{U}$ (Eq. $8$), and the decoder layer with skip attention $\Theta$(Eq. $11$). We also define the input image for the current stage as $x^t$, the Virtual Query Bank $\mathcal{B}_{vq}$, and the total training iteration $M$.
Specifically, our lazy \textbf{Query Pre-alignment} strategy is described in the line-$4$ and line-$8$-$9$, our \textbf{Consistent Selection Loss} strategy is described in the line-$5$-$7$, and our \textbf{Virtual Query} strategy is described in line-$11$-$12$, line-$10$-$16$. All model and code will be made publicly available.

\begin{algorithm}
    \caption{Pseudocode for SimCIS}
    \label{algo:stage-2}
    \begin{algorithmic}[1]
        \REQUIRE{
             Backbone $f_{\text{backbone}}$, pixel decoder $f_{\text{pixel}}$ and prototype $\mathcal{P}$ at stage $t$ and ${t-1}$; \\
             Select feature points function $\Phi$ (Eq. $4$); \\ 
             Consistent selection loss function $l_{\text{csl}}$(Eq. $6$); \\
             Calculate sample weights function $g$ (Eq. $9$); \\
             $\mathcal{B}_{vq}$ update function $\mathcal{U}$(Eq. $8$); \\
             Decoder layer with skip attention $\Theta$ (Eq. $11$); \\
             Image of current stage $x^t$;  \\
             Virtual Query Bank $\mathcal{B}_{vq}$; \\
             Training iteration $M$. \\
        }
        \ENSURE{$\mathcal{M}^t$: model of current stage. }
        \STATE $\mathcal{\sigma} \gets$ Collect pseudo-distribution statistics  
        \STATE $\mathcal{\omega} \gets$ $g(\sigma)$
        \FOR{$i \gets 1, \dots, M$}
            \STATE $F^t \gets f_{\text{pixel}}(f_{\text{encoder}}(x^t))$
            \STATE $F^{t-1} \gets f_{\text{pixel}}^{t-1}(f_{\text{encoder}}^{t-1}(x^t))$
            \STATE $\mathcal{I}^{t-1} \gets \Phi(F^{t-1}, \mathcal{P}^{t-1})$
            \STATE $\mathcal{L}_{csl} \gets l_{\text{cls}}(F^t, F^{t-1}, \mathcal{I}^{t-1}, \mathcal{P}^{t-1}) ~\triangleright$ Sec.~$4.2$ end.
            \STATE $\mathcal{I}^t \gets \Phi(F^t, P^t)$
            \STATE $Q_N \gets$ Object query on $F^t$ by $\mathcal{I}^t$. $\triangleright$ Sec.~$4.1$ end.
            \STATE $Q_j \gets$ Sample $j$ virtual query from $\mathcal{B}_{vq}$ using $\mathcal{\omega}$.
            \STATE $Q_{N+j} \gets$ $\{Q_N, Q_j\}$
            \FOR{$l \gets 1, \dots, L$}
                \STATE $Q_{N+j} \gets \Theta(Q_{N+j})$
            \ENDFOR
            \STATE $Z_N \gets$ Get $Q_N$'s prediction results.
            \STATE $\mathcal{B}_{vq} \gets \mathcal{U}(Z_N, Q_N, y)$ $\quad \triangleright$ Sec.~$4.3$ end.
            \STATE Calculate $L_{\text{class}}$ using $Q_{N+j}$.
            \STATE Calculate $L_{\text{mask}}$ using $Q_N$.
            \STATE $L_{\text{total}} \gets L_{\text{class}} + L_{\text{mask}} + L_{\text{csl}}$
            \STATE Update parameters via backpropagation.
        \ENDFOR
    \end{algorithmic}
    \label{AL:1}
    % \vspace{10pt}
\end{algorithm}

\section{More Dataset and Implementation Details}
\label{sec:suppl_imp}
\textbf{Dataset Information.} 
Following previous works~\cite{(eclipse)kim2024eclipse,(comformer)cermelli2023comformer,(CoMasTRe)gong2024continual}, we use ADE20k~\cite{(ade20k)zhou2017scene} to train and evaluate our model for both continual panoptic segmentation and continual semantic segmentation tasks. The ADE20K dataset contains $20,210$ training images and $2,000$ validation images, with each image averaging $19.5$ instances and $10.5$ classes. Compared with other datasets, such as VOC~\cite{(voc)everingham2010pascal}, which contains an average of $2.3$ instances and $1.4$ classes per image. ADE20K is a particularly challenging dataset that highlights our robustness during continual training stages.

\begin{table*}[t]
\centering
  \begin{adjustbox}{max width=0.95\linewidth}
    \begin{tabular}{c|ccc|ccc}
        \toprule
         \multirow{2}*{Random ID} & \multicolumn{3}{c}{Our SimCIS} & \multicolumn{3}{c}{ECLIPSE} \\
         & \textit{1-100}~~ & ~~\textit{101-150} & \textit{all} & \textit{1-100}~~ & ~~\textit{101-150} & \textit{all} \\
        \midrule
        
         1~~~& 41.2 & 28.9 & 37.1 & 33.4 & 20.4 &	29.1  \\
         2~~~& 42.2 & 30.2 & 38.2 & 32.1 & 23.0 & 29.1\\
         
         3~~~& 41.1 & 29.8 & 37.3 & 32.2 &	23.3 & 29.3 \\
         
         4~~~& 42.2 & 29.6 & 38.0 & 30.4 & 18.0 & 26.3 \\
         
         5~~~& 41.2 & 30.5 & 37.6 & 32.2 & 22.8 & 29.1 \\
         
         6~~~& 41.7 & 27.5 & 37.0 & 28.5 & 24.3 & 27.1 \\
         
         7~~~& 41.9 & 28.8 & 37.6 & 34.3 & 18.8 & 29.2 \\
         
         8~~~& 40.0 & 29.9 & 36.6 & 30.4 & 22.7 & 27.9 \\
         
         9~~~& 42.0 & 28.7 & 37.6 & 32.7 & 22.2 & 29.2 \\
         
         10\dag& 39.1 & 33.8 & 37.4 & 11.3 & 0.0 & 7.6 \\
         \midrule
         Origin & 42.2 & 30.1 & 38.1 & 41.4 & 18.8 &	33.9 \\
         \bottomrule
    \end{tabular}
    \end{adjustbox}
    \caption{\textbf{Continual Panoptic Segmentation with 10 random order}  on the ADE20K 100-5 continual panoptic segmentation scenario. \dag{ }means descending order. Origin means original ascending order.}
    \label{tab:suppl_random}
\end{table*}

\noindent\textbf{Implementation Details. }
To ensure a fair comparison, we strictly follow previous works~\cite{(eclipse)kim2024eclipse,(comformer)cermelli2023comformer, (L2P)wang2022learning, (MiB)cermelli2020modeling}. In the initial training step, the learning rate is set up to $1$e-$4$, and during the incremental learning phase, it is reduced to $5e$-$5$. The total training iteration is set to $160,000$ in the first step and $ 1,000$  iterations for each class in incremental steps. We utilize a multi-step strategy to dynamically adjust our learning rate for optimizing our model, with a decay factor set to $0.1$.  
Following~\cite{(MiB)cermelli2020modeling}, there are two different experimental protocols: disjoint and overlap. In the disjoint setting, each task has its own exclusive image data, while the overlap setting allows different images to appear across tasks. We choose the more challenging overlap setting as our experimental protocol. Except for setting consistent select loss weight to $2.0$, we follow Mask2former~\cite{(mask2former)cheng2022masked} to set other loss weights.

\section{Continual Learning with Random Order}
\label{sec:suppl_order}
\noindent \textbf{Experiment Details.}
As shown in Tab.~\ref{tab:suppl_random}, we conduct extensive experiments on our model and ECLIPSE~\cite{(eclipse)kim2024eclipse} under the ten random orders (detailed orders shown in Tab.~\ref{tab:rebut_order}), where nine of them were completely randomly generated using the \texttt{random} module in \texttt{Numpy} without any manual selection. As ADE20k's classes are ranked by their total pixel ratios in the entire dataset, we deliberately set the last order to descending to evaluate the model's dependency on base categories. Specifically, the descending order forces the model first to learn rare categories, enabling us to assess its continual learning ability under such challenging conditions.

\noindent \textbf{Comparison with ECLIPSE.}
The results are shown in Tab.~\ref{tab:suppl_random}. Our model achieves SOTAs across all 10 random orders. Overall, our model achieves an increase of $41.9\%$ across all classes compared to ECLIPSE. Specifically, the average performance of old classes improves by $+11.5\%$ PQ, and new classes see an average improvement of $+10.2\%$ PQ. In the final experiment, where we set the categories in descending order, the performance of ECLIPSE is relatively dropped by $73.9\%$. This demonstrates that ECLIPSE's approach, which freezes other parameters and employs the VPT \cite{(vpt)jia2022visual} strategy for model updates, strongly depends on the base class during continual learning. In contrast, our model remains stable even under this highly challenging setup.

\section{More Ablation Study for Stop Gradient}
\label{sec:suppl_stopg}
As we mention in the main text, we apply stop gradient on selected object query \( Q_N \) after the QPA strategy, to ensure that the information in feature map \( F \) is not disrupted during training, keeping the objectness information stable across different stages. As shown in the Tab.~\ref{tab:detach}. After using the stop gradient strategy, we achieve an increase of +2.1\% PQ  across all classes. All the experiments in the main text use this strategy unless otherwise specified.

\begin{table}[h]
  \centering
  \vspace{5pt}
      \begin{adjustbox}{max width=0.9\linewidth}
      \begin{tabular}{c|c|c|c|c|ccc}
        \toprule
        \multirow{2}{*}{Psd} & \multirow{2}{*}{QPA} & \multirow{2}{*}{CSL} & \multirow{2}{*}{VQ} &\multirow{2}{*}{SG} & \multicolumn{3}{c}{\textbf{Panoptic 100-5} (11 tasks)}\\
         &  &   &  && \textit{1-100} & \textit{101-150} & \textit{all}\\
        \midrule
         \checkmark & \checkmark & \checkmark & \checkmark && 39.5 & 20.7 & 33.3\\
         \checkmark & \checkmark & \checkmark & \checkmark & \checkmark & 42.1 & 21.9 & 35.4\\
        \bottomrule
      \end{tabular}
      \end{adjustbox}
      \caption{
        \textbf{Ablation Study on Stop Gradient.} Psd: pseudo label, QPA: lazy query pre-alignment, CSL: consistent selection loss, and SG: stop gradient.
      }
      \label{tab:detach}
      \vspace{-22pt}
\end{table}

\section{More Visualization Results for CSS}
\label{sec:suppl:css}
\begin{figure*}[t]
    \centering
    \includegraphics[width=0.9\linewidth]{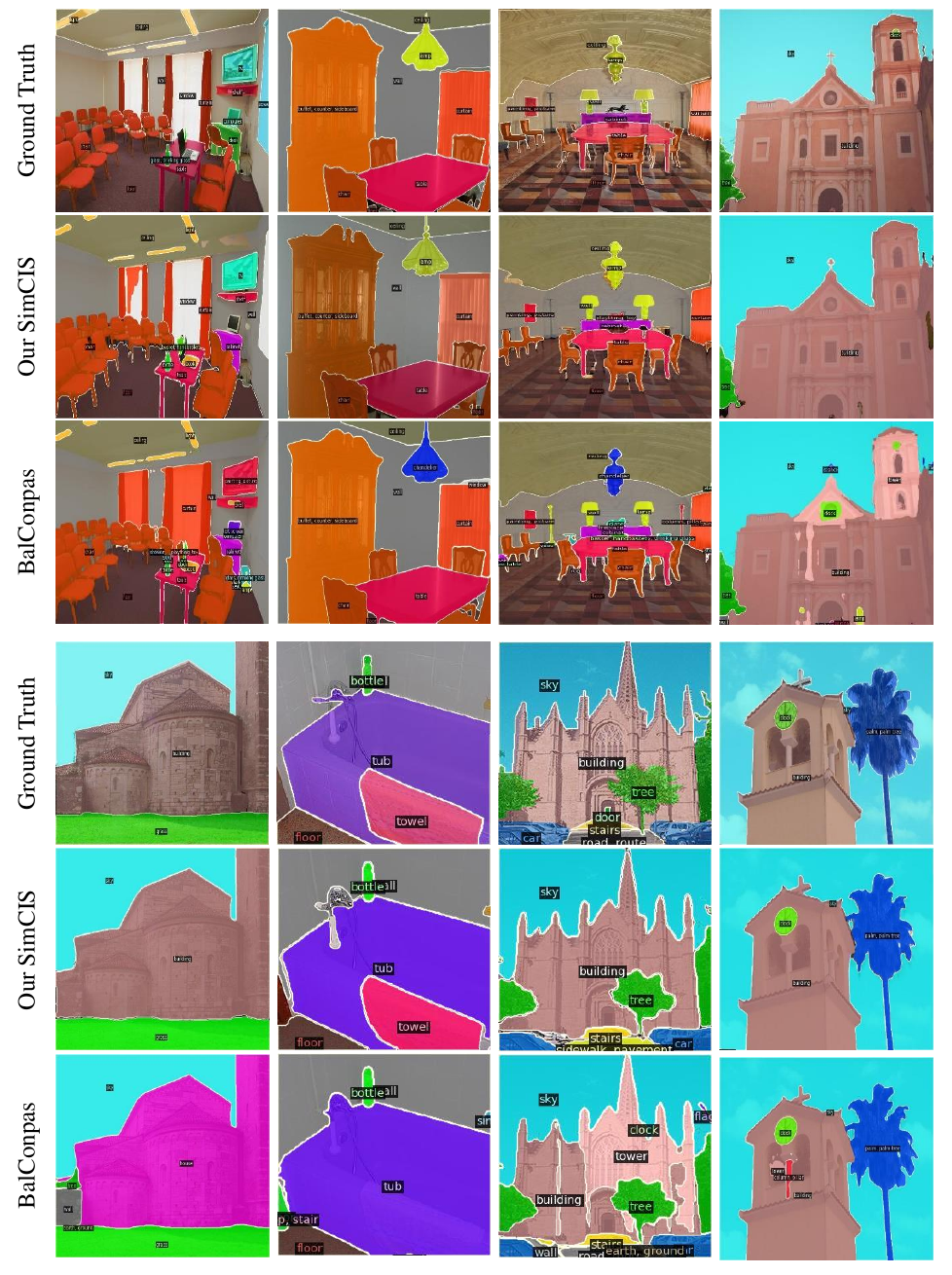}
    \caption{
        \textbf{Qualitative comparisons} between SimCIS and BalConpas~\cite{(bal)chen2024bal} on the ADE20K 100-5 continual semantic segmentation.}
    \label{fig:suppl_css}
\end{figure*}

As shown in Fig.~\ref{fig:suppl_css}, we additionally compare our SimCIS with BalConpas \cite{(bal)chen2024bal} in the 100-5 continual semantic segmentation task. In the first, second, and fourth row from Fig.~\ref{fig:suppl_css}, BalConpas encounters misclassification of the TV and lamps. In the fourth image, Balconpas fails to predict the building's accurate mask. While benefiting from the proper utilization of semantic priors in pixel feature and VQ strategy's ability to preserve class information, our SimCIS performs well in these cases.
\section{Built-in Objectness Maintenance}
\label{sec:suppl_kmeans}

\begin{figure*}
    \centering
    \includegraphics[width=0.85\linewidth]{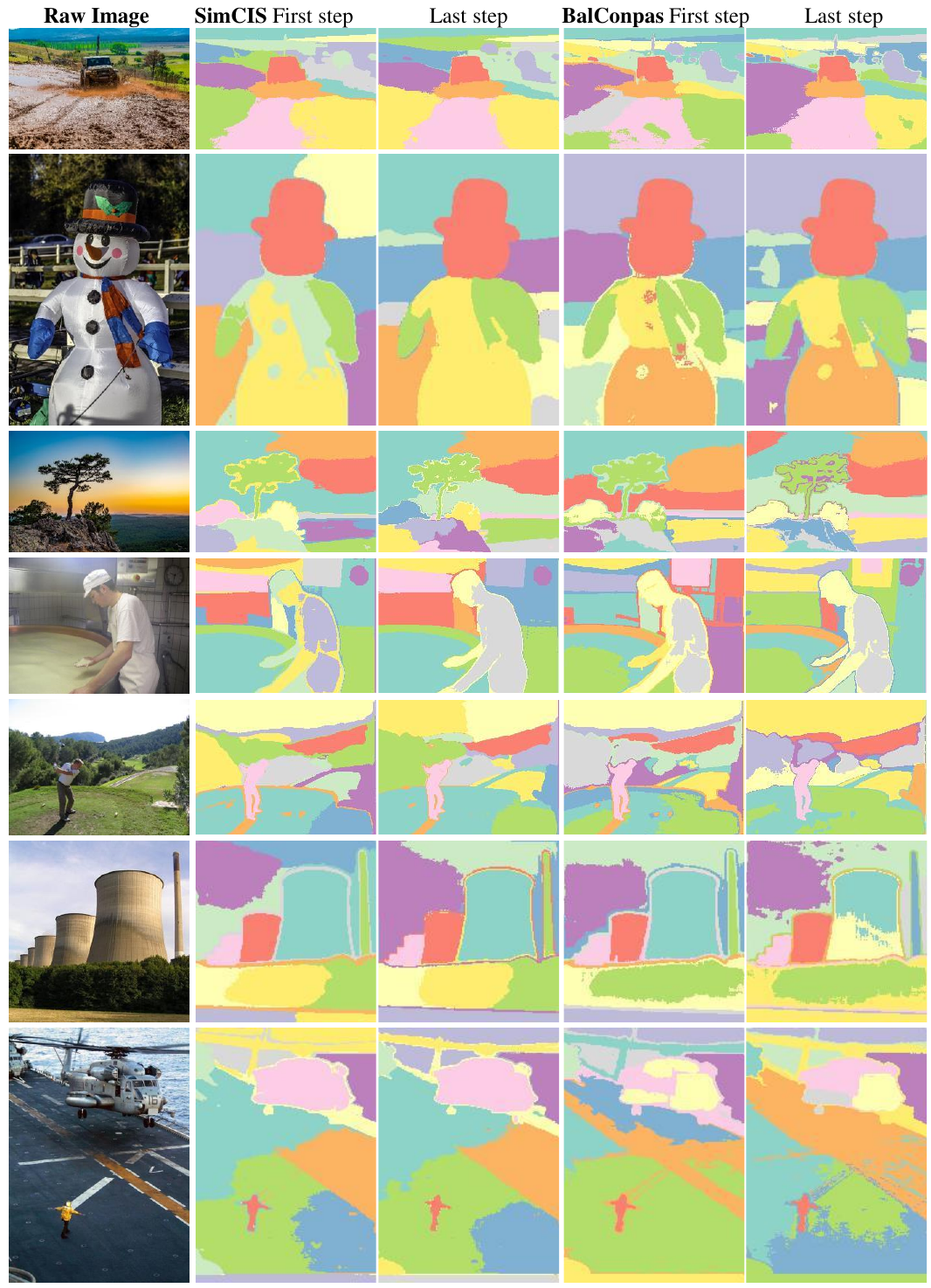}
    \caption{\textbf{Clustering results comparison between SimCIS and BalConpas.} Our SimCIS maintains the semantic priors in the pixel feature.}
    \label{fig:suppl_kmeans}
\end{figure*}

\noindent \textbf{Detailed clustering implementation.}
In the multi-scale feature generated by the pixel decoder, we choose the feature with the highest resolution for clustering. To evaluate the quality of objectness information contained in the features, we applied the K-means~\cite{hartigan1975clustering} algorithm for clustering. Regarding the hyperparameter settings, for the images shown in Fig.~\ref{fig:suppl_kmeans}, we set the number of clustering centers from top to bottom as $[15, 10, 15,15,15,15,15]$.

\noindent \textbf{SimCIS provides stable built-in objectness.}
Although pixel features can generally provide semantic priors across various methods, our observations indicate that they are still influenced by the continual learning process. In this section, we visually demonstrate that our SimCIS has the ability to maintain object information.
As shown in Fig.~\ref{fig:suppl_kmeans}, in the first image, the clustering results of Balconpas around the jeep exhibit significantly more noise. In the last image, Balconpas fails to capture the entire helicopter, while our feature successfully preserves the complete object information.

\section{The Order of Attention Layers}
In Mask2Former~\cite{(mask2former)cheng2022masked}, the authors employ a cross then self-attention mechanism, as they argue that query features to the first self-attention layer are image-independent and do not have signals from the image, thus applying self-attention is unlikely to enrich information. However, in our proposed Lazy Query Pre-alignment strategy, the query features have rich information. Therefore, we revert to the conventional sequence of cross then self-attention. This modification, however, does not exhibit any significant impact on the experimental outcomes.

\section{Discussion, Limitation and Future Work}
\label{sec:suppl_discussion}
\noindent \textbf{Discussion of the choice of meta-architecture for image segmentation.} 
To ensure a fair comparison, we adopt the same Mask2Former~\cite{(mask2former)cheng2022masked} as our meta-architecture for image segmentation. 
However, recent years have witnessed rapid advancements in transformer-based universal image segmentor~\cite{(mask-dino)li2023mask,(oneformer)jain2023oneformer}, which achieves a much stronger performance on the segmentation benchmark. We leave the investigation of other meta-architectures as future work.

\noindent \textbf{Discussion of other common techniques/tricks in CIS.}
To maintain the simplicity and elegance of our SimCIS, we have discarded certain continual learning techniques/tricks commonly used in previous methods, such as model weight fusion across stages~\cite{(ewf)xiao2023ewf}, specific initialization methods~\cite{xie2025early, cha2021ssul, baek2022decomposed} for the classifier head, and freezing model parameters~\cite{(eclipse)kim2024eclipse,(CoMasTRe)gong2024continual}. Whether these techniques/tricks can further improve SimCIS's performance remains an open question for future work.

\begin{table*}[th]
\small
\centering
  \begin{adjustbox}{max width=0.9\linewidth}
    \begin{tabularx}{\textwidth}{c|X}
        \toprule
         \multirow{1}*{ID} & \multicolumn{1}{c}{Category Order} \\
        \midrule
         1~~~& 
         \textit{[71, 135, 3, 60, 74, 1, 10, 40, 118, 91, 52, 50, 59, 146, 33, 42, 66, 148, 41, 78, 46, 14, 26, 57, 73, 96, 89, 55, 149, 84, 13, 2, 77, 54, 32, 138, 64, 81, 129, 104, 93, 86, 62, 130, 21, 125, 128, 136, 12, 65, 79, 43, 4, 134, 68, 145, 99, 15, 58, 29, 111, 51, 56, 11, 117, 102, 140, 105, 116, 131, 18, 120, 22, 19, 85, 28, 0, 123, 38, 95, 115, 17, 70, 61, 20, 112, 109, 67, 98, 133, 30, 76, 49, 8, 101, 47, 25, 48, 147, 132, 100, 44, 69, 6, 53, 126, 7, 75, 90, 83, 107, 106, 9, 113, 37, 122, 121, 143, 103, 137, 80, 144, 94, 142, 110, 63, 124, 87, 35, 24, 88, 39, 139, 27, 92, 23, 114, 119, 141, 108, 5, 45, 72, 31, 36, 127, 82, 16, 97, 34]}\\
        \midrule
         
         2~~~& \textit{[11, 114, 103, 122, 48, 41, 85, 92, 113, 64, 3, 80, 110, 10, 112, 30, 96, 101, 102, 9, 7, 21, 17, 37, 93, 77, 73, 94, 59, 135, 2, 123, 98, 130, 49, 129, 25, 66, 50, 145, 76, 147, 83, 90, 63, 111, 27, 126, 1, 65, 75, 119, 12, 78, 5, 143, 15, 29, 71, 22, 89, 115, 84, 16, 120, 139, 38, 68, 146, 116, 35, 124, 97, 23, 39, 117, 13, 18, 108, 138, 33, 134, 141, 62, 105, 142, 40, 26, 8, 46, 144, 95, 131, 99, 104, 19, 60, 132, 6, 42, 4, 140, 128, 55, 32, 70, 118, 100, 125, 127, 87, 52, 45, 31, 81, 88, 44, 24, 20, 56, 82, 61, 28, 34, 148, 14, 53, 121, 47, 133, 57, 137, 67, 136, 106, 36, 58, 109, 107, 72, 91, 86, 43, 74, 69, 0, 149, 51, 79, 54]}\\
           \midrule
         3~~~& \textit{[74, 149, 75, 46, 113, 67, 118, 89, 130, 7, 119, 33, 77, 39, 96, 81, 112, 37, 124, 1, 34, 105, 35, 80, 135, 13, 143, 53, 9, 101, 22, 57, 139, 138, 12, 123, 48, 63, 60, 69, 117, 71, 4, 65, 127, 84, 97, 59, 70, 91, 128, 142, 41, 99, 136, 32, 108, 120, 42, 145, 148, 104, 87, 132, 52, 5, 85, 61, 10, 121, 49, 44, 17, 115, 93, 134, 68, 3, 110, 36, 133, 102, 0, 16, 55, 90, 83, 54, 62, 94, 126, 6, 19, 18, 26, 51, 114, 31, 43, 45, 76, 131, 25, 66, 92, 29, 50, 40, 100, 58, 109, 20, 30, 98, 86, 14, 28, 107, 122, 11, 111, 64, 21, 72, 103, 137, 23, 88, 125, 140, 47, 146, 27, 116, 141, 78, 79, 24, 95, 2, 144, 38, 82, 56, 106, 129, 147, 73, 8, 15]} \\
           \midrule
         4~~~& \textit{[60, 110, 89, 119, 147, 123, 116, 35, 22, 1, 36, 99, 58, 17, 43, 11, 109, 130, 113, 138, 65, 94, 74, 8, 106, 12, 29, 118, 24, 136, 140, 21, 6, 93, 142, 9, 71, 135, 54, 114, 121, 77, 16, 105, 117, 5, 67, 86, 61, 97, 20, 76, 18, 84, 103, 46, 96, 0, 141, 100, 63, 131, 31, 45, 81, 73, 13, 124, 79, 48, 40, 132, 102, 112, 107, 44, 27, 49, 134, 85, 144, 66, 83, 104, 75, 88, 101, 82, 19, 47, 87, 122, 125, 115, 72, 137, 7, 128, 78, 15, 90, 51, 145, 39, 2, 126, 64, 139, 41, 55, 34, 26, 3, 129, 69, 68, 120, 98, 92, 57, 59, 70, 23, 80, 148, 10, 149, 52, 38, 42, 53, 108, 127, 91, 50, 95, 146, 56, 33, 30, 111, 25, 62, 32, 4, 37, 14, 143, 133, 28]} \\
           \midrule
         5~~~& \textit{[77, 20, 111, 65, 117, 53, 43, 90, 28, 79, 134, 45, 116, 98, 92, 105, 137, 10, 6, 59, 67, 34, 44, 99, 55, 147, 1, 80, 122, 54, 56, 12, 31, 49, 37, 61, 108, 133, 143, 130, 70, 95, 132, 2, 115, 118, 81, 47, 51, 121, 14, 3, 8, 21, 22, 62, 78, 72, 39, 25, 23, 142, 149, 50, 83, 11, 52, 141, 129, 113, 4, 148, 144, 136, 91, 146, 35, 114, 46, 138, 97, 16, 69, 84, 131, 64, 66, 5, 24, 13, 68, 9, 102, 104, 139, 106, 74, 126, 19, 0, 58, 60, 96, 32, 41, 94, 7, 48, 93, 30, 119, 75, 42, 15, 57, 38, 127, 120, 124, 100, 135, 123, 63, 33, 103, 71, 128, 17, 145, 26, 86, 29, 107, 82, 88, 73, 110, 112, 85, 89, 27, 125, 109, 40, 76, 87, 36, 101, 18, 140]} \\
           \midrule
         6~~~& \textit{[54, 27, 42, 13, 38, 94, 134, 97, 95, 109, 130, 26, 117, 67, 107, 96, 69, 78, 141, 113, 4, 147, 129, 108, 144, 145, 49, 44, 128, 115, 148, 104, 19, 58, 114, 89, 98, 21, 106, 39, 138, 63, 43, 7, 12, 17, 81, 84, 103, 45, 120, 5, 23, 142, 143, 14, 102, 56, 116, 112, 136, 60, 50, 92, 65, 82, 127, 139, 8, 91, 10, 93, 131, 83, 73, 74, 85, 75, 121, 105, 40, 25, 123, 149, 118, 52, 29, 88, 126, 51, 110, 1, 122, 133, 47, 99, 137, 80, 55, 57, 62, 71, 125, 140, 32, 20, 2, 61, 132, 30, 111, 37, 76, 64, 15, 77, 79, 28, 33, 100, 31, 124, 72, 119, 9, 6, 90, 36, 16, 68, 22, 59, 86, 18, 0, 70, 53, 3, 34, 41, 46, 35, 24, 135, 146, 101, 66, 87, 11, 48]}\\
           \midrule
         7~~~& \textit{[87, 70, 74, 1, 60, 111, 0, 26, 59, 35, 57, 128, 55, 24, 20, 53, 108, 49, 140, 29, 54, 6, 84, 10, 101, 5, 94, 32, 79, 63, 15, 9, 31, 107, 110, 104, 38, 33, 77, 132, 43, 149, 72, 119, 37, 56, 112, 114, 124, 13, 51, 58, 47, 83, 69, 45, 11, 145, 127, 123, 52, 97, 98, 8, 73, 95, 117, 86, 46, 89, 65, 93, 62, 61, 129, 28, 39, 125, 78, 67, 133, 120, 14, 99, 21, 141, 121, 7, 136, 42, 88, 17, 146, 19, 131, 96, 102, 4, 34, 44, 30, 22, 50, 90, 142, 137, 81, 82, 16, 118, 130, 100, 103, 64, 18, 113, 135, 41, 12, 85, 2, 115, 147, 134, 80, 76, 66, 68, 36, 109, 3, 105, 106, 92, 75, 138, 148, 27, 126, 71, 40, 48, 25, 139, 91, 122, 116, 23, 143, 144]}\\
           \midrule
         8~~~& \textit{[22, 119, 103, 67, 40, 38, 95, 43, 72, 34, 54, 88, 132, 94, 0, 107, 91, 104, 71, 21, 133, 16, 1, 27, 48, 125, 139, 144, 35, 75, 129, 25, 53, 82, 117, 7, 140, 124, 128, 147, 120, 23, 70, 122, 108, 106, 93, 12, 90, 73, 149, 99, 52, 47, 146, 28, 61, 55, 37, 87, 76, 136, 112, 148, 29, 57, 49, 45, 65, 100, 13, 32, 68, 78, 58, 69, 56, 2, 9, 130, 110, 51, 116, 123, 111, 118, 101, 19, 138, 59, 109, 4, 85, 98, 17, 141, 131, 50, 92, 8, 81, 30, 6, 41, 79, 97, 46, 74, 126, 115, 31, 11, 15, 3, 33, 5, 63, 105, 83, 62, 64, 134, 39, 137, 113, 36, 42, 10, 18, 114, 145, 80, 84, 66, 60, 77, 86, 89, 14, 127, 24, 96, 121, 142, 20, 143, 26, 44, 135, 102]}\\
           \midrule
         9~~~&\textit{[83, 53, 93, 75, 14, 89, 54, 2, 115, 80, 110, 24, 56, 124, 62, 113, 1, 30, 100, 107, 86, 82, 87, 95, 129, 149, 0, 130, 143, 103, 43, 122, 29, 106, 19, 34, 5, 17, 74, 90, 6, 97, 44, 139, 51, 31, 35, 135, 96, 9, 72, 18, 66, 33, 40, 126, 125, 91, 23, 145, 94, 77, 3, 78, 49, 27, 7, 50, 63, 28, 41, 55, 84, 73, 123, 42, 38, 8, 102, 109, 112, 119, 65, 121, 144, 88, 133, 132, 25, 114, 134, 105, 92, 10, 11, 120, 79, 26, 47, 16, 46, 137, 71, 141, 117, 48, 20, 101, 142, 15, 104, 21, 127, 136, 147, 140, 128, 32, 108, 70, 57, 98, 69, 45, 22, 111, 12, 99, 59, 60, 36, 52, 116, 58, 13, 68, 76, 4, 131, 146, 67, 39, 148, 37, 138, 64, 118, 85, 61, 81]}\\
           \midrule
         10*& \textit{[149, 148, 147, 146, 145, 144, 143, 142, 141, 140, 139, 138, 137, 136, 135, 134, 133, 132, 131, 130, 129, 128, 127, 126, 125, 124, 123, 122, 121, 120, 119, 118, 117, 116, 115, 114, 113, 112, 111, 110, 109, 108, 107, 106, 105, 104, 103, 102, 101, 100, 99, 98, 97, 96, 95, 94, 93, 92, 91, 90, 89, 88, 87, 86, 85, 84, 83, 82, 81, 80, 79, 78, 77, 76, 75, 74, 73, 72, 71, 70, 69, 68, 67, 66, 65, 64, 63, 62, 61, 60, 59, 58, 57, 56, 55, 54, 53, 52, 51, 50, 49, 48, 47, 46, 45, 44, 43, 42, 41, 40, 39, 38, 37, 36, 35, 34, 33, 32, 31, 30, 29, 28, 27, 26, 25, 24, 23, 22, 21, 20, 19, 18, 17, 16, 15, 14, 13, 12, 11, 10, 9, 8, 7, 6, 5, 4, 3, 2, 1, 0]}\\
         \bottomrule
    \end{tabularx}
    \end{adjustbox}
    \caption{\textbf{Random orders.}}
    \label{tab:rebut_order}
    \vspace{10mm}
\end{table*}

\end{document}